\pdfoutput=1
\documentclass[letterpaper,10 pt]{ieeeconf}
\usepackage[letterpaper,left=19.1mm,right=19.1mm,top=19.1mm,bottom=19.1mm]{geometry}
\usepackage{ifpdf}
\usepackage{cite}
\usepackage[cmex10]{amsmath}
 \usepackage{algorithm}   
 \usepackage{algpseudocode}
\usepackage{array}
\usepackage{mdwmath}
\usepackage{mdwtab}
\usepackage{eqparbox}
\usepackage{booktabs}
\usepackage[colorlinks,linkcolor=black, urlcolor=black]{hyperref}
\usepackage[caption=false,font=footnotesize]{subfig}
 \usepackage{mathptmx} 
 \usepackage{times} 
 \usepackage{amsmath} 
 \usepackage{amssymb}  %
 \usepackage{ifpdf}
 \usepackage{graphics} 
 \usepackage{epsfig} 
 \usepackage{mathptmx} 
 \usepackage{times} 
 \usepackage{amsmath} 
 \usepackage{amssymb}  %
 \usepackage{algorithm}   
 \usepackage{algpseudocode}
\usepackage{mathrsfs}
\usepackage{mathtools}
 \usepackage{array}
 \usepackage{mdwmath}
 \usepackage{mdwtab}
 \usepackage{eqparbox}
 \usepackage{stackengine}
\usepackage{gensymb}
\usepackage{fixltx2e}
\usepackage{stfloats}
 \DeclareMathAlphabet{\mathpzc}{OT1}{pzc}{m}{it}

\usepackage{url}
\usepackage{multirow,array}
\usepackage{mwe}
\usepackage{graphicx}
\graphicspath{{./results/}}
 \DeclareGraphicsExtensions{.pdf,.jpeg,.png,.jpg}   
\IEEEoverridecommandlockouts                              
    \usepackage{algorithm}
    \usepackage{algpseudocode}
\usepackage{graphics} 
\usepackage{epsfig} 
\usepackage{mathptmx} 
\usepackage{times} 
\usepackage{amsmath} 
\usepackage{amssymb}  %

\newcommand{\etal}{\textit{et al.}}

\title{\textbf{Semantic Reinforced Attention Learning for Visual Place Recognition}}

\author
{Guohao Peng$^1$, Yufeng Yue$^2$, Jun Zhang$^1$, Zhenyu Wu$^1$, Xiaoyu Tang$^1$ and Danwei Wang$^1$,~\IEEEmembership{Fellow,~IEEE}
\thanks{$^1$G. Peng, J. Zhang, Z. Wu, X. Tang and D. Wang are with School of Electrical and Electronic Engineering, Nanyang Technological University, 639798, Singapore (email: peng0086@ntu.edu.sg)}
\thanks{$^2$Y. Yue is with the School of Automation, Beijing Institute of Technology, Beijing 100081, China}
}

\begin{document}
\maketitle
\thispagestyle{empty}
\pagestyle{plain}
\begin{abstract}
Large-scale visual place recognition (VPR) is inherently challenging because not all visual cues in the image are beneficial to the task. 
In order to highlight the task-relevant visual cues in the feature embedding,
the existing attention mechanisms are either based on artificial rules or trained in a thorough data-driven manner.
To fill the gap between the two types,
we propose a novel Semantic Reinforced Attention Learning Network (SRALNet), in which the inferred attention can benefit from both semantic priors and data-driven fine-tuning.
The contribution lies in two-folds.
 (1) To suppress misleading local features, an interpretable local weighting scheme is proposed based on hierarchical feature distribution.
 (2) By exploiting the interpretability of the local weighting scheme, a semantic constrained initialization is proposed so that the local attention can 
 be reinforced by semantic priors.
 Experiments demonstrate that our method outperforms state-of-the-art techniques on city-scale VPR benchmark datasets.
\end{abstract}

\IEEEpeerreviewmaketitle
\section{Introduction}

Visual place recognition  (VPR) has been a crucial research field in computer vision~\cite{Arandjelovic2016NetVLADCA,Arandjelovic2014DisLocationSD,Cao2013GraphBasedDL,Schindler2007CityScaleLR,Sattler2015HyperpointsAF,Torii2015247PR} and robotics~\cite{Chen2011CityscaleLI,Cummins2008FABMAPPL,wu2019magnetic,DBLP:journals/ijrr/CumminsN11,Merrill2018LightweightUD} communities, 
since it is the cornerstone of many popular
applications 
including
autonomous driving~\cite{Chalmers2018LearningTP,McManus2014ShadyDR,MurArtal2015ORBSLAMAV}, geo-localization~\cite{Laskar2017CameraRB,Sattler2012ImageRF,Lim2012RealtimeI6,wu2021mstsl} and 3D reconstruction~\cite{Crandall2011DiscretecontinuousOF}.

Typically, VPR is
solved as an image retrieval task~\cite{Sattler2012ImageRF,Arandjelovic2014DisLocationSD,Torii2015247PR,Arandjelovic2016NetVLADCA,Kim2017LearnedCF,Khaliq2018AHV},
where 
the most similar reference images 
are retrieved
when given a query image.
City-scale  VPR 
has always been challenging, because even the same scene may undergo great appearance changes due to different weather, illumination, and viewpoints.
Partial occlusion and dynamic objects 
will also increase the task difficulty.
Therefore,
how to form a robust image representation
has become the focus of research in the field.

Among all attempts to construct compact and powerful image representations 
in the past decades,
aggregation-based methods  
have proven to be fruitful. Typical representatives range from Fisher Vector (FV)~\cite{Perronnin2010LargescaleIR}, Vector  of Locally  Aggregated  Descriptors (VLAD)~\cite{Babenko2015AggregatingLD} to recent Convnet architectures that introduce multiple pooling  strategies~\cite{Mohedano2016BagsOL,Ong2017SiameseNO,Razavian2014VisualIR,Tolias2015ParticularOR,Kalantidis2015CrossdimensionalWF}. However, not all visual cues in the image are related to the task. Early methods quantify all local features indiscriminately,
which may result in
misleading information being encoded into the image representation.

\begin{figure}[!t] 
\begin{center} 
  \includegraphics[width=0.95\linewidth]{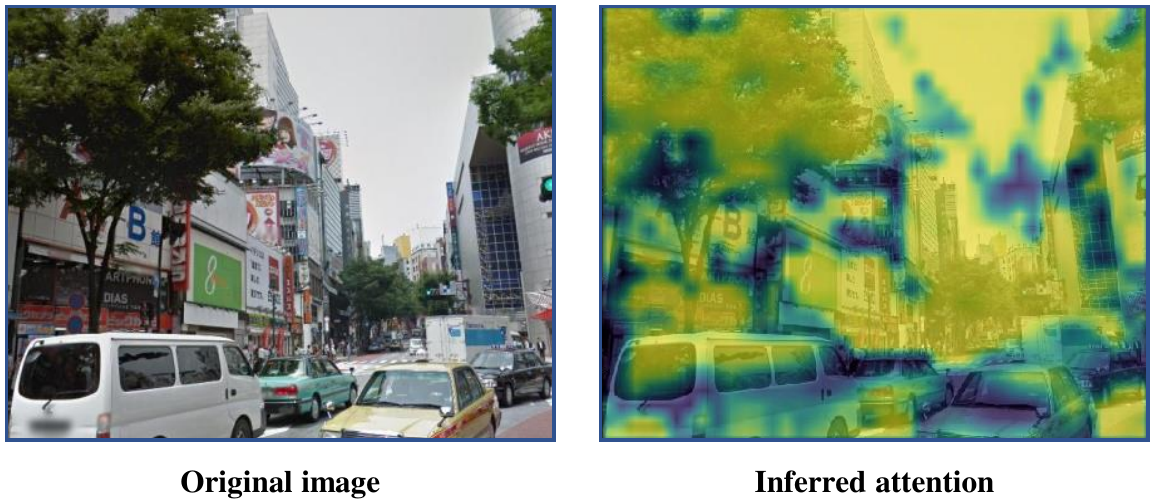} 
  \caption{
  The superimposed heat map illustrates which visual cues our model learns to suppress in feature embedding.
  As can be seen,
  the inferred attention is consistent with the human cognition 
that values long-term static structures and ignores unreliable objects.
Going beyond 
the rough prior knowledge of 
"preserving building semantics", our model learns to adaptively retain billboards and suppress repeated structures on buildings.
}\label{FIG_vis_att_1} 
\end{center}   
\vspace{-13pt}
\end{figure}

To address this problem, attention mechanism is introduced to emphasize the task-relevant local features~\cite{Knopp2010AvoidingCF,Arandjelovic2014DisLocationSD,Sattler2015HyperpointsAF}. Recent attention-aware methods can be categorized as either data-driven or rule-based.
The data-driven methods~\cite{Zhu2018AttentionbasedPA,Noh2016LargeScaleIR,Kim2017LearnedCF} integrate attention modules into an end-to-end encoding network for unified learning. 
However, they simply employ attention mechanism as the black box weighting of local features, which 
lacks the interpretability to reflect priors.
The rule-based methods typically use
semantic information to filter specific visual cues~\cite{Piasco2019LearningSG,Mousavian2015SemanticallyGL,Naseer2017SemanticsawareVL}, while their performance is limited by 
prior knowledge and the generalization ability of the semantic segmentation algorithm.
To narrow the gap between the two types, it's necessary to set up an interpretable attention module that can elegantly combine prior knowledge and data-driven learning, so that the attention can benefit from their complementary.

With this motivation, we propose an end-to-end architecture that integrates attention learned from both semantic priors and data-driven training.
Specifically,
the model incorporates an interpretable local weighting scheme which is constructed 
based on hierarchical feature distribution. 
Its intrinsic relationship with encoding space partition enables initial attention to be provided by prior knowledge.
On this basis, a semantic constrained initialization is proposed,
which equivalently provides better initial attention for the local weighting scheme.
Through further fine-tuning, the ultimate local attention can benefit from the mutual promotion between semantic priors and data-driven learning.
In accordance with the above statement, our contributions can be elaborated as follows:
\begin{itemize}
     \item 
     A novel attentional encoding architecture SRALNet is proposed for city-scale VPR, which incorporates comprehensive attention into feature embedding.
     \item An interpretable local weighting scheme({LW}) is proposed to refine local features.
     In the field of VPR,
     to our best knowledge,
     this is the first attention mechanism
     that can
     integratge semantic priors and data-driven learning.
     \item A semantic constrained initialization({SC}) is proposed to reinforce the local weighting scheme.
     It paves the way to reflect semantic priors with initial weights.
     \item Experiments are conducted to verify the effectiveness of the proposed components, and our method outperforms SOTA techniques on all benchmark datasets.
 \end{itemize}
    \begin{figure*}[!tb] 
    \begin{center} 
    	     \includegraphics[width=0.95\textwidth]{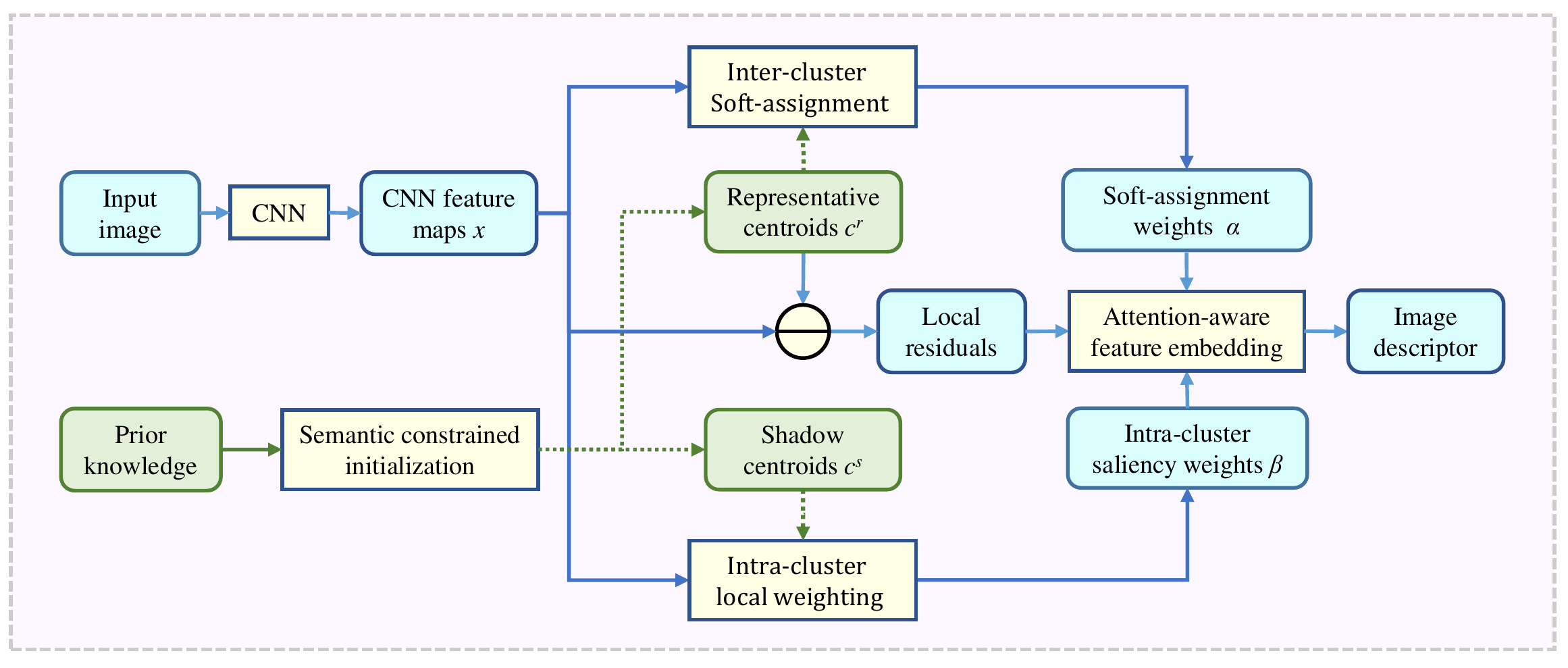}
        	 \caption{
        	 The overall flowchart of the SRALNet encoding pipeline. 
        	 The deep local features are clustered, refined, and encoded into the final image descriptor (the blue solid arrow).
        	 Semantic priors are introduced to enhance the hierarchical weighting module (the green dotted arrow).
        	 }\label{FIG_overallflowchart}
    \end{center}  
    \vspace{-12pt}
    \end{figure*}
    
\pagestyle{empty}
\section{Related Work}
Early methodologies for VPR
    count on hand-crafted local features~\cite{Dorst2011DistinctiveIF,Bay2008SpeededUpRF}
    coupled with variant BoW~\cite{Torii2015247PR,Schindler2007CityScaleLR,Sattler2015HyperpointsAF}  models.
    Later on, 
    VLAD~\cite{Babenko2015AggregatingLD} and FV~\cite{Perronnin2010LargescaleIR} have emerged as powerful alternatives.
    With the popularity of deep learning, 
    Max pooling~\cite{Razavian2014ABF}, sum pooling~\cite{Babenko2015AggregatingLD},
    VLAD~\cite{Ng2015ExploitingLF} and FV~\cite{Uricchio2015FisherEC} all show convincing advantages when 
    encoding deep local features into compact image representations.
    Studies have also demonstrated that incorporating 
    spatial information of pyramid patches~\cite{8700608} or selective regions~\cite{Snderhauf2015PlaceRW,Khaliq2018AHV}
     can boost performance. 
    A typical case is R-MAC~\cite{Tolias2015ParticularOR} that aggregates the max pooled activations of multi-scale grids.
    
    The recent
    researches~\cite{Arandjelovic2016NetVLADCA,Gordo2016DeepIR,Radenovi2016CNNIR} have shown that network fine-tuning  on a task-specific dataset can significantly improve the performance.
    Arandjelovic \etal~\cite{Arandjelovic2016NetVLADCA} propose a generalized VLAD pooling layer named NetVLAD, which is differentiable for end-to-end training. 
    Concurrently, Gordo \etal~\cite{Gordo2016DeepIR} fine-tune the R-MAC~\cite{Tolias2015ParticularOR} to fit for a large-scale landmark dataset.
    Yu \etal~\cite{8700608} propose SPENetVLAD, 
    which enhances NetVLAD by encoding the spatial information in the stacked regional features.
    These methods recruit all the local features in feature embedding,
    which may result in misleading visual cues degrading the image representation.

    To selectively embed task-relevant local features, Kim \etal~\cite{Kim2017LearnedCF} extend the NetVLAD~\cite{Arandjelovic2016NetVLADCA} with a contextual weighting network(CRN), which 
    utilizes the information from
    features'
    context.
    Zhu \etal~\cite{Zhu2018AttentionbasedPA} propose APAnet that integrates a cascaded attention scheme before pyramid aggregation of local features. 
    More recently, high-level prior knowledge has proven conducible for scene classification and place recognition. In ~\cite{Piasco2019LearningSG,Mousavian2015SemanticallyGL,Naseer2017SemanticsawareVL}, semantic information is employed as supervision to harvest local features from man-made discriminative visual objects for the subsequent embedding.
    Noticeably, these attention modules are either artificial rule-based or harnessed with black-box weighting. 
    In this paper, we manage to combine prior knowledge and data-driven learning in 
    an interpretable local weighting scheme.

    Outside the field of VPR, 
    the most related work is GhostVLAD~\cite{Zhong2018GhostVLADFS} proposed for face recognition.
    With the same architecture as NetVLAD, it specifies the clusters for characterizing vague faces and excludes them from feature embedding. 
    Adapting GhostVLAD to the VPR tasks  requires specific initialization.
    It is also imperfect to exclude or retain an entire cluster from the embedding,
    since local features from the same cluster cannot be all misleading or informative.
    SRALNet decouples clustering and filtering through a hierarchical weighting scheme, where more flexible refinement is performed within each cluster.

\section{Semantic Reinforced Attention Learning}\label{shaodow}
In order to highlight the divergent importance of visual cues to the task,
we propose an attention-aware encoding architecture named SRALNet.
Fig.\ref{FIG_overallflowchart} presents the overall flowchart of the encoding pipeline.
The following subsections will describe how we introduce local attention to suppress the misleading visual cues in the image representation.

\subsection{Deep Local Feature Clustering} 
\begin{figure*}
         \centering
	    \includegraphics[width=1\textwidth]{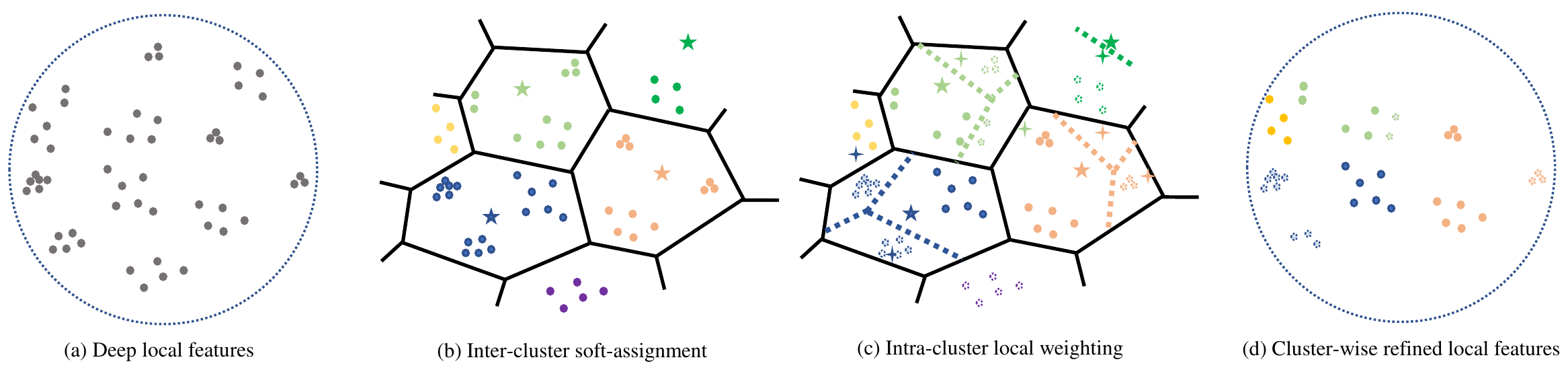}
    	 \caption{
    	 The illustration of the hierarchical weighting within SRALNet. 
    	 (a)$\sim$(c) show how deep local features are clustered and refined through the inter-cluster soft-assignment and intra-cluster local weighting.
        (d) visualizes the local features 
        that are 
        reserved for the 
        subsequent
        feature embedding after the hierarchical weighting. 
    	 }\label{FIG_pipeline}
    	 \vspace{-12pt}
\end{figure*}

Following the common local feature representation\cite{Arandjelovic2016NetVLADCA,Kim2017LearnedCF,Zhu2018AttentionbasedPA}, we exploit a deep convolutional neuron network VGG-16\cite{Simonyan2014VeryDC}
as the backbone for local feature extraction. Spatial activations ${m} \in {R^{D \times 1 \times 1}}$ decomposed from the feature maps of the last convolutional layer ${M} \in {R^{D \times H \times W}}$ are treated as deep local features.
Illustrated as
Fig.\ref{FIG_pipeline}.a, the normalized local features ${x} \in {R^{D}}$ are scattered on the unit hypersphere.

After local feature extraction,
we then adopt the soft-assignment~\cite{Arandjelovic2016NetVLADCA} to divide the local features into $K$ visual word clusters (Fig.\ref{FIG_pipeline}.b).
Defined as in Eq.(\ref{eq3}),
the soft-assignment weight $\alpha_k(\mathbf{x_i})$ 
indicates the probability of a local feature $\mathbf{x}_{i}$ belonging to the $k^{th}$ cluster.
$\mathbf{c}_k^r$ denotes the representative centroid ($\star$ in Fig.\ref{FIG_pipeline}.b) of the $k^{th}$ cluster.
The
constant $a$
controls the decay of the response with the 
magnitude of the 
distance,
which is initialized by a large positive number.
\begin{align}
    {\alpha _k}({\mathbf{x}_i})
    =\frac{e^{-a\|\mathbf{x}_i-c^r_k\|^2}}{\sum_{j=1}^K{e^{-a\|\mathbf{x}_i-c^r_j\|^2}}}
    \label{eq3}
\end{align}

\subsection{Intra-cluster Local Weighting}
To further suppress task-irrelevant features in each cluster, 
we propose a local weighting scheme based on the intra-cluster feature distribution.
As visual cues that describe similar semantics and appearance usually have consistent task relevance and are mapped to adjacent locations in the feature space,
we hypothesize the Voronoi cell of a cluster can be separated into an informative area and multiple ambiguous areas (Fig.\ref{FIG_pipeline}.c). 
Each ambiguous area is represented by a shadow centroid $\mathbf{c}_{kl}^s$ ($+$ in Fig.\ref{FIG_pipeline}.c). 
The intra-cluster saliency weight $\beta_k(\mathbf{x}_i)$ is defined as the probability of a local feature $\mathbf{x}_i$ from the $k^{th}$ cluster being located in the informative area $I$. Assuming that sub-clusters are uniformly distributed and each one conforms to a Gaussian with equal covariance matrix, $\beta_k(\mathbf{x}_i)$ can be derived through the Bayesian theorem as in Eq.(\ref{eq4}):
\begin{equation}
\begin{aligned}
    {\beta _k}({\mathbf{x}_i})&=P(I|\mathbf{x}_i,c_k)\\
    &=\frac{P(\mathbf{x}_i|I,c_k)P(I|c_k)}{\sum_{l=1}^S{P(\mathbf{x}_i|s_l,c_k)P(s_l|c_k)}+P(\mathbf{x}_i|I,c_k)P(I|c_k)}\\
    &=\frac{e^{-a\|\mathbf{x}_i-\mathbf{c}_k^r\|^2}}{\sum_{l=1}^S{e^{-a\|\mathbf{x}_i-\mathbf{c}_{kl}^s\|^2}}+e^{-a\|\mathbf{x}_i-\mathbf{c}_k^r\|^2}}
    \label{eq4}
\end{aligned}
\end{equation}

According to Eq.(\ref{eq4}), 
local features 
located in
ambiguous areas
will be assigned with a low saliency weight. 
Although a total of $S$ shadow centroids are initialized for each cluster, some of them could be cast away from the cluster boundary after optimization and become ineffective.
This provides a flexible internal partition for each cluster.
Essentially, the local weighting acts as the re-allocation of local features into sub-clusters where informative area is highlighted and ambiguous ones are suppressed. 
This interpretability makes it possible to introduce prior attention by initializing the encoding space partition, 
which is intrinsically determined by the centroids $\mathbf{c}_k^r$ and $\mathbf{c}_k^s$ according to Eq.(\ref{eq3}) and Eq.(\ref{eq4}).
This characteristic
will be further leveraged in Section.\ref{semantic}.

\subsection{Unified Implementation of The Hierarchical Weighting}
It can be noticed that by expanding the square terms,
both Eq.(\ref{eq3}) and Eq.(\ref{eq4}) can be simplified by canceling out 
$e^{-a\|\mathbf{x}_i\|^2}$ 
between the numerator and the
denominator. 
Additionally, by using
the abbreviated symbol $\mathbf{c}_{kn}$ to represent $\mathbf{c}_k^r$ and $\{\mathbf{c}_{kl}^s\}$ in the $k^{th}$ cluster,
where $n$=0 denotes the representative centroid and the others denote the shadow ones, 
$\alpha_k(\mathbf{x}_i)$ and $\beta_k(\mathbf{x}_i)$ can be further derived as in Eq.(\ref{eq1})
and Eq.(\ref{eq100}). Since both transforms take local features as inputs, they can be implemented through a unified convolutional layer, followed by the Softmax function across the specified channels. 
\begin{align}
    &{\alpha _k}({\mathbf{x_i}})=\frac{e^{2a{\mathbf{c}_{k0}^\mathrm{T}\mathbf{x_i}}-a{{\| \mathbf{c}_{k0} \|}^2}}}{\sum_{j=1}^K{e^{2a{\mathbf{c}_{j0}^\mathrm{T}\mathbf{x_i}}-a{{\| \mathbf{c}_{j0} \|}^2}}}}    =\frac{e^{{\mathbf{w}_{k0}^\mathrm{T}\mathbf{x}_i}+b_{k0}}}{\sum_{j=1}^K{e^{{\mathbf{w}_{j0}^\mathrm{T}\mathbf{x}_i}+b_{j0}}}}\label{eq1}\\
    &{\beta _k}({\mathbf{x}_i})=\frac{e^{2a{\mathbf{c}_{k0}^\mathrm{T}\mathbf{x}_i}-a{{\| \mathbf{c}_{k0} \|}^2}}}{\sum_{l=0}^S{e^{2a{\mathbf{c}_{kl}^\mathrm{T}\mathbf{x}_i}-a{{\| \mathbf{c}_{kl} \|}^2}}}}
    =\frac{e^{{\mathbf{w}_{k0}^\mathrm{T}\mathbf{x}_i}+b_{k0}}}{\sum_{l=0}^S{e^{{\mathbf{w}_{kl}^\mathrm{T}\mathbf{x}_i}+b_{kl}}}}\label{eq100}
\end{align}


\subsection{Attention-aware Image Representation}
After the division and refinement of local features, 
the $k^{th}$ visual word vector $V_k$ can be calculated as a spatial aggregation of the double-weighted local residuals.
\begin{align}
    {\mathbf{V_k}} = \sum\limits_{i = 1}^{HW} {{\alpha _k}({\mathbf{x}_i})\beta_k ({\mathbf{x}_i})({\mathbf{x}_i} - \mathbf{c}_k^r)}
    \label{eq9}
\end{align}

Intuitively,
$\beta(\mathbf{x}_i)$ scales down the residual norm of misleading local features in the ambiguous areas, thereby suppressing them in the feature embedding.

\begin{figure}[htb]
         \centering
	     \includegraphics[width=0.98\linewidth]{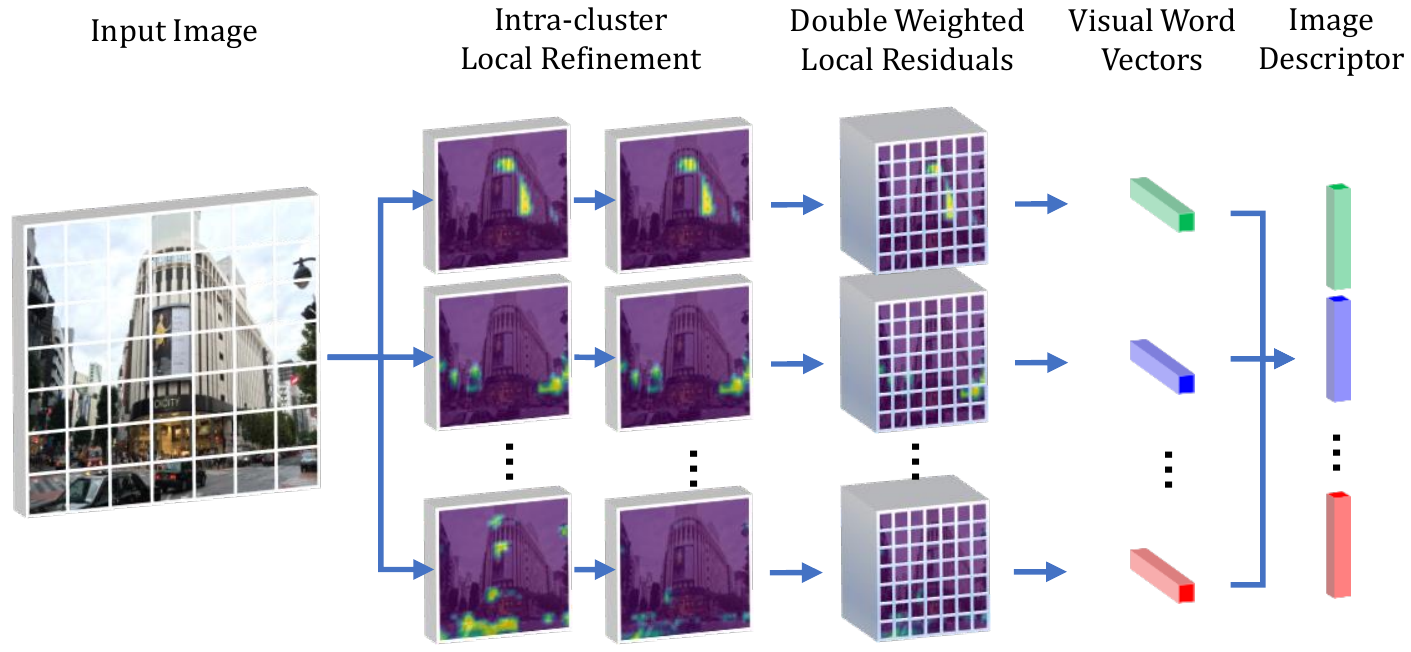}
    	 \caption{ 
    	 The diagram of the attention-aware descriptor embedding.
    	 }\label{FIG_flowchart2}
    	 \vspace{-6pt}
\end{figure}

As illustrated in Fig.\ref{FIG_flowchart2},
the final image descriptor is the concatenation of $K$ visual word vectors
    followed by intra-normalization\cite{Jgou2010AggregatingLD} and $L_2$-normalization.
To achieve the most discriminative representation,
the optimization of trainable parameters $\{\mathbf{W} \}$, $\{b\}$ and $\{\mathbf{c}_k^r\}$ in Eq.(\ref{eq1})$\sim$(\ref{eq9}) is intrinsically equivalent to explore the optimal hierarchical encoding space partition and feature allocation.

\subsection{Semantic Constrained Initialization}\label{semantic}
According to
Eq.(\ref{eq1})$\sim$(\ref{eq9}),
the initial parametric model is determined by 
the representative centroids and shadow centroids.
Thus 
a normal initialization could be achieved by mimicking traditional VLAD
as \cite{Arandjelovic2016NetVLADCA} does:
the representative centroids $\{\mathbf{c}_k^r\}$ are initialized by the clustering centers of the sampled deep local features,
while shadow centroids $\{\mathbf{c}_{kl}^s\}$ are randomly initialized to be distant from their corresponding $\mathbf{c}_k^r$.
However, not all sampled local features are task-relevant. Consequently, some of the representative centroids may be initialized by misleading visual cues.
Therefore, we introduce semantics as the prior constraints to select specific local features for initializing $\{\mathbf{c}_k^r\}$ and $\{\mathbf{c}_{kl}^s\}$ respectively.

Specifically, we adopt the 
DeepLabV3 \cite{Chen2018EncoderDecoderWA} pre-trained on Cityscapes dataset \cite{Cordts2016Cityscapes} to provide common semantic classes under urban driving scenes. The activations before Softmax prediction are first scaled to the same size of our feature map through max pooling. Then Softmax is implemented to predict the labels of local features. Features labeled as static objects, including 'building','road','traffic signs' and 'vegetation', are filtered and sampled for generating $K$ representative centroids. While those dynamic or task-irrelevant semantics, such as 'sky','person' and 'vehicle', are used for generating $N$ shadow candidates. For each cluster, the $S$ shadow centroids are initialized by the top $S$ candidates that have the closest Euclidean distances with the representative centroid.

\begin{figure}[!t]
         \centering
	     \includegraphics[width=0.95\linewidth]{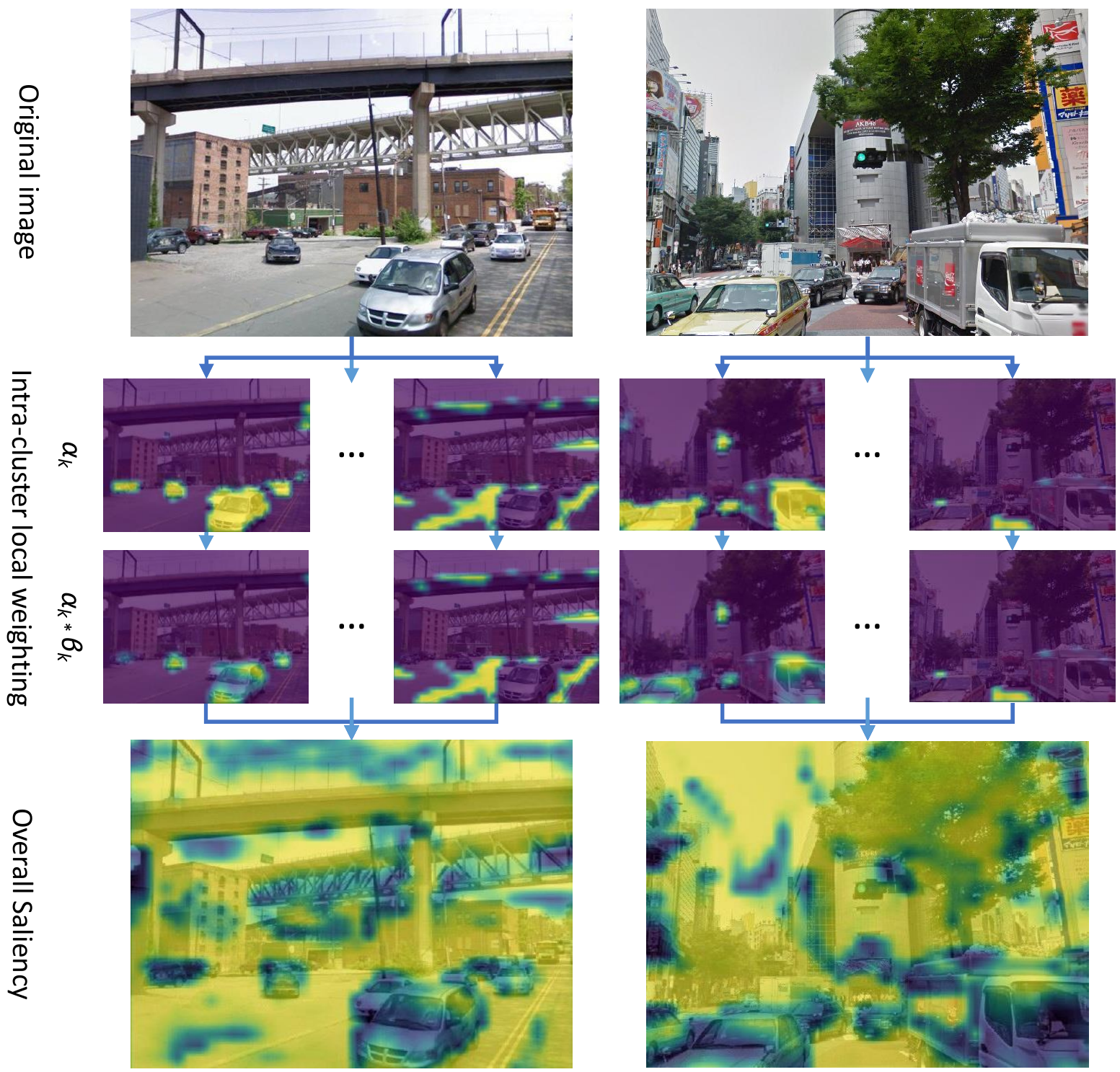}
    	 \caption{ 
    	 Attention inference of two examples from Pittsburgh and TokyoTM. The second row shows the cluster-wise saliency of visual cues before and after intra-cluster weighting, from which it can be observed that the vehicles are suppressed and the road markings are preserved.
    	 }\label{FIG_vis_sc}
    	 \vspace{-6pt}
\end{figure}

The semantic constrained initialization essentially partitions the encoding space based on semantic priors, which equivalently
provides better initial attention for the local weighting scheme. On this basis, we allow the network to fit the optimal attention through end-to-end training.
Thereby, the ultimate local attention can benefit from the mutual promotion between semantic priors and data-driven learning. 
As illustrated in Fig.\ref{FIG_vis_att_1} and Fig.\ref{FIG_vis_sc},
although no precise pixel-level semantic annotations are required 
for supervision,
the learned attention still turns out to be largely consistent with human cognition that inhibits task-irrelevant semantics.

    \section{Training Pipeline}\label{train}
    Let $f_\theta$ be the image representation and $\theta$ be the parameters to be trained.
    To make a positive reference $I_r^p$ closer to the query $I_q$ than any negative candidate $I_r^n$ in feature space,
    we adopt the triplet ranking loss~\cite{Gordo2016EndtoEndLO,7298682,8545492,Radenovi2016CNNIR,9072271} as in Eq.(\ref{eq17}) for metric learning. 
    \begin{equation}
    \begin{aligned}
        l_\theta(I_q,I_r^p,I_r^n) = &[d^2(f_\theta(I_q),f_\theta(I_r^p))\\
        &-d^2(f_\theta(I_q),f_\theta(I_r^n))+m]_+
        \label{eq17}
    \end{aligned}  
    \end{equation}
    
    $[x]_+=max(x,0)$ 
    and $m$ denotes the empirical margin.
    We follow the same positive and hard negative mining in \cite{Arandjelovic2016NetVLADCA} to prepare a set of tuples $(I_q,I_r^{p*},\{I_r^n\})$ for training.
    A tuple consists of 
    1 query, 1 positive and $N$ negatives, which can be further divided into $N$ triplets $(I_q,I_r^{p*},I_r^{nj})$. The loss for each tuple can be expressed as:
    \begin{align}
        L_\theta(I_q,I_r^{p*},\{I_r^n\}) = \frac{1}{N}\sum\limits_{j = 1}^N{l_\theta(I_q,I_r^{p*},I_r^{nj})}
        \label{eq18}
    \end{align}
    
    By minimizing Eq.(\ref{eq18}), the parametric model is trained under an weakly supervised manner, with only GPS tags used for threshold-based tuple mining for query images. 
    
    \section{Experiments}\label{experiments}
    This section describe how we conduct experiments to 
    validate our proposed method.
    
\begin{table*}[!tb]
	\caption{Evaluation of the proposed components (\textbf{LW} and \textbf{SC}). The comparison is made with other generalized VLAD pooling layers. All representations are based on VGG-16 architecture without PCA whitening.} 
	\centering
	\label{table_2}
    \begin{tabular}{l|c|c|c c c|c c c|c c c}
      \hline
      \multirow{2}*{Method}&\multirow{2}*{LW}&\multirow{2}*{SC}
      &\multicolumn{3}{c|}{Pitts30k-test} &\multicolumn{3}{c|}{Pitts250k-test}&\multicolumn{3}{c}{Tokyo24/7}\\
      \cline{4-12}
        &{}&{}&r@1&r@5&r@10&r@1&r@5&r@10&r@1&r@5&r@10\\
      \hline
      NetVLAD~\cite{Arandjelovic2016NetVLADCA}&{$\times$}&{$\times$}&83.6&92.2&94.0&84.1&92.5&94.5 &60.0&76.2&79.7\\
      {GhostVLAD~\cite{Zhong2018GhostVLADFS}}&{$\surd$}&{$\times$}&83.7&92.5&94.7&84.1&92.7&95.1&61.6&76.5&80.6\\
      {CRN~\cite{Kim2017LearnedCF}}&{$\surd$}&{$\times$}&84.0&92.6&94.9&84.7&92.9&95.3 &61.9&75.6&79.7\\
      {Ours: SRALNet}&{$\surd$}&{$\times$}&84.4&92.5&94.8&84.8&93.1&95.4&63.5&77.8&81.0\\
      {{Ours: SRALNet-SC}}&{$\surd$}&{$\surd$}&\textbf{85.1}&\textbf{93.3}&\textbf{95.2}&\textbf{85.8}&\textbf{94.1}&\textbf{95.9}&\textbf{68.6}&\textbf{80.0}&\textbf{83.8}\\
      \hline
	\end{tabular}
\end{table*}    
\begin{figure}[t]
         \centering
	     \includegraphics[width=8cm]{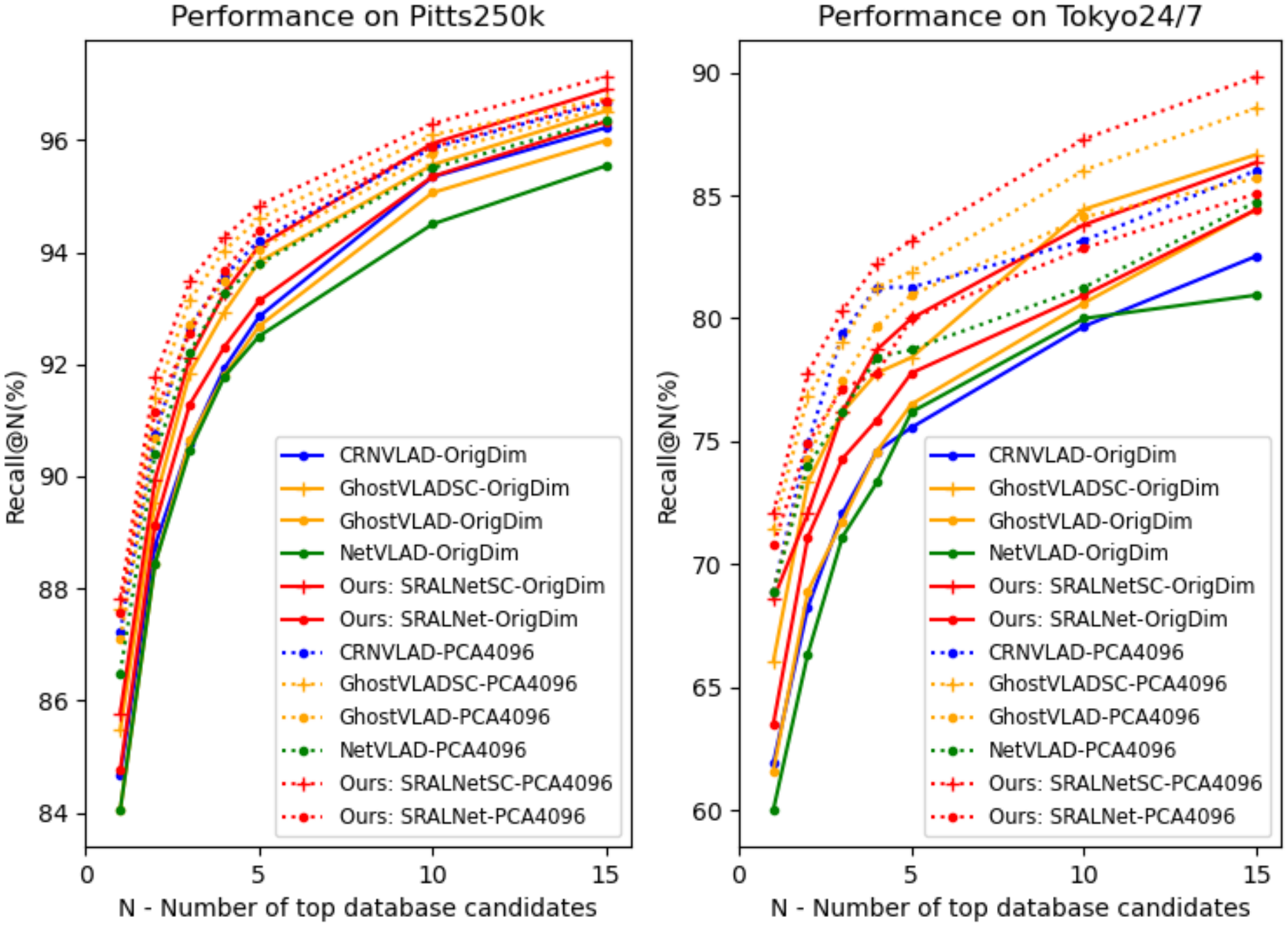}
    	 \caption{ 
    	 Representations applying \textbf{SC}(-+-) steadily outperform those without \textbf{SC}(-$\cdot$-). Performing \textbf{PCA-W} to reduce original dimensionality(--) to 4096-D($\cdot$$\cdot$$\cdot$) improves the performance of all models. Our SRALNet(red) surpasses all corresponding counterparts(yellow, blue, green) in all cases.
    	 }\label{FIG_eval_sc}
\end{figure}

    \subsection{Datasets and Evaluation Metric}\label{EM}
    In this work, three benchmark datasets 
    of 
    retrieval-based VPR\cite{Arandjelovic2016NetVLADCA,Kim2017LearnedCF,Liu2018StochasticAE,Zhu2018AttentionbasedPA}
    are employed for the evaluation.
    \textbf{Pitts250k}~\cite{Torii2013VisualPR} contains 254k images captured in Pittsburgh area,  which are geographically divided into three subsets for training, validation and testing.
    \textbf{Pitts30k}~\cite{Arandjelovic2016NetVLADCA} is a refined subset of Pitts250k, with 
    train/val/test sets all 
    containing a 10k database and around 7k queries.
    \textbf{Tokyo 24/7}~\cite{Arandjelovic2016NetVLADCA,Torii2015247PR} is a 
    more challenging dataset, which
    contains 76k database images and 315 query images captured at daytime, sunset and night.
    Same as the latest SOTA~\cite{8700608},
    we employ Pitts30k-train as the only training set, and test the trained models on Pitts250k-test, Pitts30k-test, Tokyo24/7 respectively.
    
    Following
    the standard evaluation protocol~\cite{Arandjelovic2016NetVLADCA,Kim2017LearnedCF,Zhu2018AttentionbasedPA},
    the performance of a retrieval inference is measured by the recall given $N$ potential positive candidates ($Recall@N$). 
    The indexing of a query is deemed successful as long as one of the candidates falls within a distance of $d_r$=25$m$ from the geographic location of the query image. 

    \subsection{Benchmark Methods}
    Five state-of-the-art models proposed for VPR
    are selected for comparison:
    \textbf{NetVLAD}~\cite{Arandjelovic2016NetVLADCA} is the seminal generalized VLAD pooling layer.
    On top of it,
    \textbf{CRN}~\cite{Kim2017LearnedCF} introduces an attention layer to estimate local saliency according to semi-global context. 
    \textbf{SPENetVLAD}~\cite{8700608} 
    stacks the 
    regional 
    VLAD
    features to retain the spatial information.
    \textbf{R-MAC}~\cite{Tolias2015ParticularOR} 
    aggregates 
    the max-pooled activations of multi-scale rigid grids.  
    \textbf{APAnet}~\cite{Zhu2018AttentionbasedPA} 
    aggregates
    spatial pyramid features 
    weighted by cascaded attention blocks. \
    Besides, we adapt 
     \textbf{GhostVLAD}~\cite{Zhong2018GhostVLADFS}, a similar architecture proposed for face recognition, to the VPR tasks. 
\setlength{\tabcolsep}{4pt}
\begin{table}[!t]
\renewcommand\arraystretch{1.0}
	\caption{Evaluate the separate contribution of semantic constrained initialization (\textbf{SC}) and data-driven fine-tuning (\textbf{FT}). 
	}
	\centering
	\label{table_1.5}
    \begin{tabular}{l|c c|c c c|c c c}
      \hline
      \multirow{2}*{\footnotesize{Method}}&\multirow{2}*{\footnotesize{SC}}&\multirow{2}*{\footnotesize{FT}}  &\multicolumn{3}{c|}{\footnotesize{Pitts250k-test}}&\multicolumn{3}{c}{\footnotesize{Tokyo24/7}}\\
      \cline{4-9}
      &&&\footnotesize{r}\footnotesize{@1}&\footnotesize{r}\footnotesize{@5}&\footnotesize{r}\footnotesize{@10}&\footnotesize{r}\footnotesize{@1}&\footnotesize{r}\footnotesize{@5}&\footnotesize{r}\footnotesize{@10}\\
      \hline
      \multirow{2}*{\footnotesize{NetVLAD~\cite{Arandjelovic2016NetVLADCA}}}&{-}&{$\times$}&\footnotesize{68.2}&\footnotesize{81.7}&\footnotesize{85.6}&\footnotesize{42.9}&\footnotesize{62.9}&\footnotesize{69.8}\\
      &{-}&{$\surd$}&\textbf{\footnotesize{84.1}}&\textbf{\footnotesize{92.5}}&\textbf{\footnotesize{94.5}}&\textbf{\footnotesize{60.0}}&\textbf{\footnotesize{76.2}}&\textbf{\footnotesize{79.7}}\\
      \hline
      \multirow{4}*{\footnotesize{GhostVLAD~\cite{Zhong2018GhostVLADFS}}}&{$\times$}&{$\times$}&\footnotesize{68.2}&\footnotesize{81.7}&\footnotesize{85.5}&\footnotesize{44.1}&\footnotesize{62.9}&\footnotesize{70.8}\\
      &{$\surd$}&{$\times$}&\footnotesize{71.7}&\footnotesize{84.6}&\footnotesize{88.3}&\footnotesize{54.0}&\footnotesize{69.2}&\footnotesize{74.6}\\
      &{$\times$}&{$\surd$}&\footnotesize{84.1}&\footnotesize{92.7}&\footnotesize{95.1}&\footnotesize{61.6}&\footnotesize{76.5}&\footnotesize{80.6}\\
      &{$\surd$}&{$\surd$}&\textbf{\footnotesize{85.5}}&\textbf{\footnotesize{93.8}}&\textbf{\footnotesize{95.6}}&\textbf{\footnotesize{66.0}}&\textbf{\footnotesize{78.4}}&\textbf{\footnotesize{84.4}}\\
      \hline
      \multirow{4}*{\footnotesize{Ours: SRALNet}}&{$\times$}&{$\times$}&\footnotesize{68.7}&\footnotesize{82.2}&\footnotesize{86.0}&\footnotesize{43.2}&\footnotesize{64.1}&\footnotesize{70.8}\\
      &{$\surd$}&{$\times$}&\footnotesize{72.6}&\footnotesize{85.5}&\footnotesize{89.2}&\footnotesize{56.5}&\footnotesize{70.5}&\footnotesize{76.8}\\
      &{$\times$}&{$\surd$}&\footnotesize{84.8}&\footnotesize{93.1}&\footnotesize{95.4}&\footnotesize{63.5}&\footnotesize{77.8}&\footnotesize{81.0}\\
      &{$\surd$}&{$\surd$}&\textbf{\footnotesize{85.8}}&\textbf{\footnotesize{94.1}}&\textbf{\footnotesize{95.9}}&\textbf{\footnotesize{68.6}}&\textbf{\footnotesize{80.0}}&\textbf{\footnotesize{83.8}}\\
      \hline
	\end{tabular}
\end{table}
    \subsection{Implementation Details}
    Since different deep learning frameworks were used in the previous SOTAs~\cite{Arandjelovic2016NetVLADCA,Kim2017LearnedCF,Zhu2018AttentionbasedPA,8700608}, we re-implement all comparative models in {Pytorch} 
    for fair comparison.
    The pre-trained VGG-16\cite{Simonyan2014VeryDC} 
    cropped at the last convolutional layer is employed as the 
    base network for local feature extraction.
    Benchmark models and ours are implemented as a subsequent pooling layer appended to the base network. 
    The number of visual words $K$ in evaluated VLAD variants is uniformly set to 64.
    All models are trained and evaluated using the same pipeline.
    We use Stochastic Gradient Descent (SGD) optimizer 
    (with initial learning rate 0.01, momentum 0.9 and weight decay 0.001) 
    to minimize the loss function Eq.(\ref{eq18}), in which the margin $m$ is chosen as 0.1. 
    All models are
    trained for 30 epochs, with the learning rate reduced by a factor of 2 every 5 epochs. An early stop is triggered once the best validation $Recall@1$ stagnates for 10 epochs.
    For more compact representations,
    we also perform PCA whitening (\textbf{PCA-W}) and $L_2$-normalization 
    on the baselines and our method.

\subsection{Ablation Study}
As elaborated in Section \ref{shaodow},
comprehensive attention has been integrated into the encoding strategy through the proposed components. To validate the local weighting scheme (\textbf{LW}) and semantic constrained initialization (\textbf{SC}) respectively,
the plain SRALNet (with \textbf{LW} only) is set as the base model while \textbf{SC} is an applicable option for ablation study.

We compare the performance of our method with other generalized VLAD pooling layers.
GhostVLAD
has the same architecture as NetVLAD, but 
excludes
the specified clusters from the descriptor embedding.
SRALNet and CRN differ from NetVLAD with an additional weighting for local features: CRN 
acts as 
a black-box estimator for predicting the local saliency, while the local weighting scheme within SRALNet is interpretable which is derived from the feature distribution.
The top retrieved results of different models are given in Table \ref{table_2}. 
As can be seen, the models with local weighting mechanism
(SRALNet, CRN and GhostVLAD)
all surpass NetVLAD on both benchmarks.
One can easily judge that introducing attention in local feature refinement conduces to the discriminability of generated descriptors. 
Besides, both SRALNet and CRN outperforms plain GhostVLAD, which shows the advantages of decoupling clustering and filtering.
SRALNet surmounts CRN
in the retrieval performance, 
which indicates the superiority of our interpretable local weighting scheme.
Furthermore, applying semantic constrained initialization can bring stable performance improvements.
As shown in Table \ref{table_2} and Fig.\ref{FIG_eval_sc}, SRALNet-SC convincingly outperforms all other baselines on both benchmarks, especially for Tokyo24/7 where a great improvement of 9\% has been achieved in Recall@1 index compared with NetVLAD. 
It demonstrates that endowing the local weighting scheme with semantic attention priors brings more robustness to the embedded descriptor,
and its advantages are more pronounced in more challenging scenarios.

\setlength{\tabcolsep}{3.8pt}
\begin{table}[!tb]
	\caption{Performance comparison with other generalized VLAD pooling layers in original-D and 4096-D representations.}
	\centering
	\label{table_4}
    \begin{tabular}{l|c|c c c|c c c}
      \hline
      \multirow{2}*{\footnotesize{Method}}&\multirow{2}*{\footnotesize{PCA-W}}  &\multicolumn{3}{c|}{\footnotesize{Pitts250k-test}}&\multicolumn{3}{c}{\footnotesize{Tokyo24/7}}\\
      \cline{3-8}
      &&\footnotesize{r}\footnotesize{@1}&\footnotesize{r}\footnotesize{@5}&\footnotesize{r}\footnotesize{@10}&\footnotesize{r}\footnotesize{@1}&\footnotesize{r}\footnotesize{@5}&\footnotesize{r}\footnotesize{@10}\\
      \hline
      \multirow{2}*{\footnotesize{NetVLAD~\cite{Arandjelovic2016NetVLADCA}}}&\footnotesize{w/o}&\footnotesize{84.1}&\footnotesize{92.5}&\footnotesize{94.5}&\footnotesize{60.0}&\footnotesize{76.2}&\footnotesize{79.7}\\
      &{\scriptsize{4096D}}&\textbf{\footnotesize{86.5}}&\textbf{\footnotesize{93.8}}&\textbf{\footnotesize{95.5}}&\textbf{\footnotesize{68.9}}&\textbf{\footnotesize{78.7}}&\textbf{\footnotesize{81.3}}\\
      \hline
      \multirow{2}*{\footnotesize{GhostVLAD}~\cite{Zhong2018GhostVLADFS}}&\footnotesize{w/o}&\footnotesize{84.1}&\footnotesize{92.7}&\footnotesize{95.1}&\footnotesize{61.6}&\footnotesize{76.5}&\footnotesize{80.6}\\
      &{\scriptsize{4096D}}&\textbf{\footnotesize{87.1}}&\textbf{\footnotesize{94.1}}&\textbf{\footnotesize{95.8}}&\textbf{\footnotesize{68.9}}&\textbf{\footnotesize{81.0}}&\textbf{\footnotesize{84.1}}\\
      \hline
      \multirow{2}*{\footnotesize{CRN~\cite{Kim2017LearnedCF}}}&\footnotesize{w/o}&\footnotesize{84.7}&\footnotesize{92.9}&\footnotesize{95.3}&\footnotesize{61.9}&\footnotesize{75.6}&\footnotesize{79.7}\\
      &{\scriptsize{4096D}}&\textbf{\footnotesize{87.2}}&\textbf{\footnotesize{94.2}}&\textbf{\footnotesize{95.9}}&\textbf{\footnotesize{68.9}}&\textbf{\footnotesize{81.3}}&\textbf{\footnotesize{83.2}}\\
      \hline
      \multirow{2}*{\footnotesize{Ours: SRALNet}}&\footnotesize{w/o}&\footnotesize{84.8}&\footnotesize{93.1}&\footnotesize{95.4}&\footnotesize{63.5}&\footnotesize{77.8}&\footnotesize{81.0}\\
      &{\scriptsize{4096D}}&\textbf{\footnotesize{87.6}}&\textbf{\footnotesize{94.4}}&\textbf{\footnotesize{95.9}}&\textbf{\footnotesize{70.8}}&\textbf{\footnotesize{80.0}}&\textbf{\footnotesize{85.1}}\\
      \hline
      \multirow{2}*{\footnotesize{Ours: SRALNet-SC}}&\footnotesize{w/o}&\footnotesize{85.8}&\footnotesize{94.1}&\footnotesize{95.9}&\footnotesize{68.6}&\footnotesize{80.0}&\footnotesize{83.8}\\
      &{\scriptsize{4096D}}&\textbf{\footnotesize{87.8}}&\textbf{\footnotesize{94.8}}&\textbf{\footnotesize{96.3}}&\textbf{\footnotesize{72.1}}&\textbf{\footnotesize{83.2}}&\textbf{\footnotesize{87.3}}\\
      \hline
	\end{tabular}
\end{table}

The outstanding performance of SRALNet-SC can be attributed to the mutual promotion between semantic priors and data-driven fine-tuning.
To evaluate their contribution separately, 
we first compare the off-the-shelf models with and without semantic prior enhancement.
Then we compare their performance with and without further fine-tuning.
As can be seen in Table \ref{table_1.5}, the off-the-shelf NetVLAD, GhostVLAD and SRALNet perform similarly, it is because they are all initialized to mimic traditional VLAD.
Through semantic constrained initialization, the local features predicted to have misleading semantics are basically located in shadow areas/ghost clusters and suppressed.
Thereby, the off-the-shelf SRALNet-SC/GhostVLAD-SC can be regarded as ‘rule-based filter + VLAD’. 
The results show that introducing semantic priors for local feature filtering steadily raises the performance for both models, while further fine-tuning can bring another significant improvement. Besides, SRALNet surpasses GhostVLAD in all cases, which demonstrates the greater optimization potential of our hierarchical architecture.
\subsection{Comparison with SOTA methods} 
Table \ref{table_4} shows the comparison of our model with other generalized VLAD variants in two dimensional representations.
One can infer that
the reduction into 4096 dimensions using
PCA whitening consistently improves the retrieval performance of all evaluated architectures. It can be attributed to the fact that PCA whitening could penalize the co-occurance over-counting\cite{Jgou2012NegativeEA} while preserving the energy distribution.
Table \ref{table_5} represents the comparison of performance on a more compact representation with 512 dimensions, where our method outperforms all the baseline counterparts. Combining the results in Table \ref{table_2}$\sim$\ref{table_5}, it can be seen that our proposed SRALNet has shown compelling advantages in different dimensional representations.


\setlength{\tabcolsep}{3.8pt}
\begin{table}[t]
	\caption{Performance comparison with 512-D representations. 
	All comparative models are reimplemented in Pytorch, trained and evaluated using the same protocol as our model. 
	}
	\centering
	\label{table_5}
    \begin{tabular}{l|c|c c c|c c c}
      \hline
      \multirow{2}*{\footnotesize{Method}}&\multirow{2}*{\footnotesize{PCA-W}}  &\multicolumn{3}{c}{\footnotesize{Pitts250k-test}}&\multicolumn{3}{c}{\footnotesize{Tokyo24/7}}\\
      \cline{3-8}
      &&{r}\footnotesize{@1}&{r}\footnotesize{@5}&{r}\footnotesize{@10}&{r}\footnotesize{@1}&{r}\footnotesize{@5}&{r}\footnotesize{@10}\\
      \hline
      {\footnotesize{Max pooling} \cite{Razavian2014ABF}}&{\scriptsize{512D}}&\footnotesize{39.3}&\footnotesize{59.0}&\footnotesize{67.2}&\footnotesize{11.8}&\footnotesize{23.5}&\footnotesize{33.3}\\
      {\footnotesize{R-Mac} \cite{Tolias2015ParticularOR}}&{\scriptsize{512D}}&\footnotesize{54.7}&\footnotesize{72.8}&\footnotesize{78.9}&\footnotesize{27.9}&\footnotesize{49.2}&\footnotesize{56.8}\\
      {\footnotesize{Sum pooling} \cite{Babenko2015AggregatingLD}}&{\scriptsize{512D}}&\footnotesize{70.7}&\footnotesize{84.1}&\footnotesize{88.5}&\footnotesize{28.6}&\footnotesize{43.8}&\footnotesize{53.0}\\
      {\footnotesize{APAnet} \cite{Zhu2018AttentionbasedPA}}&{\scriptsize{512D}}&\footnotesize{76.7}&\footnotesize{88.8}&\footnotesize{91.7}&\footnotesize{51.1}&{\footnotesize{66.7}}&\footnotesize{71.1}\\
      \hline
      {\footnotesize{NetVLAD} \cite{Arandjelovic2016NetVLADCA}}&{\scriptsize{512D}}&\footnotesize{83.3}&\footnotesize{92.3}&\footnotesize{94.5}&\footnotesize{55.2}&\footnotesize{68.9}&\footnotesize{74.9}\\
      {\footnotesize{GhostVLAD}~\cite{Zhong2018GhostVLADFS}}&{\scriptsize{512D}}&\footnotesize{83.9}&\footnotesize{92.6}&\footnotesize{95.1}&\footnotesize{56.5}&\footnotesize{71.8}&\footnotesize{76.5}\\
      {\footnotesize{SPENetVLAD}~\cite{8700608}}&{\scriptsize{512D}}&\footnotesize{84.4}&\footnotesize{93.1}&\footnotesize{94.8}&\footnotesize{57.1}&\footnotesize{72.4}&\footnotesize{79.7}\\
      {\footnotesize{CRN}~\cite{Kim2017LearnedCF}}&{\scriptsize{512D}}&{\footnotesize{84.5}}&\footnotesize{92.9}&\footnotesize{95.0}&\footnotesize{59.1}&\footnotesize{73.7}&\footnotesize{76.8}\\
      \hline
      {\footnotesize{Ours: SRALNet-SC}}&{\scriptsize{512D}}&\textbf{\footnotesize{84.8}}&\textbf{\footnotesize{93.5}}&\textbf{\footnotesize{95.6}}&\textbf{\footnotesize{60.6}}&\textbf{\footnotesize{76.5}}&\textbf{\footnotesize{80.0}}\\
      \hline
	\end{tabular}
\end{table}
\section{Conclusions}\label{conclusions}
In this paper, we propose an attentional encoding architecture named SRALNet for 
VPR.
To suppress misleading visual cues in the representation, we propose an interpretable local weighting scheme that can elegantly integrate semantic priors and data-driven learning.
Experiments show that the comprehensive attention incorporated in the feature embedding can greatly enhance the image representation.
On the benchmark datasets for city-scale VPR, our SRALNet is proven to be superior to the state-of-the-art methods in different dimensional representations.


\bibliographystyle{IEEEtran}

 	\bibliography{2018_ijrr}
\end{document}